\documentclass[letterpaper]{article} 
\usepackage{aaai2026}  
\usepackage{times}  
\usepackage{helvet}  
\usepackage{courier}  
\usepackage[hyphens]{url}  
\usepackage{graphicx} 
\def\UrlFont{\rm}  
\usepackage{natbib}  
\usepackage{caption} 
\usepackage{algorithm}
\usepackage{algorithmic}
\usepackage{tcolorbox}
\usepackage{newfloat}
\usepackage{listings}

\usepackage{epsfig}
\usepackage{amsmath}
\usepackage{amssymb}
\usepackage{pifont}
\usepackage{makecell}
\usepackage{booktabs}
\usepackage{multirow}

\usepackage{silence}
\usepackage{marvosym}

\makeatletter

\usepackage{colortbl}  
\usepackage{xcolor} 

\DeclareCaptionStyle{ruled}{labelfont=normalfont,labelsep=colon,strut=off} 
\floatstyle{ruled}
\newfloat{listing}{tb}{lst}{}
\floatname{listing}{Listing}
\title{Unleashing the Potential of Large Language Models for Text-to-Image Generation through Autoregressive Representation Alignment}
\author{
    Xing Xie$\textsuperscript{\rm 1, \rm 2}\equalcontrib$,
    Jiawei Liu$\textsuperscript{\rm 1}\equalcontrib$,
    Ziyue Lin$\textsuperscript{\rm 3}$,
    Huijie Fan\textsuperscript{\rm 1}\thanks{Corresponding author},
    Zhi Han\textsuperscript{\rm 1},
    Yandong Tang\textsuperscript{\rm 1},
    Liangqiong Qu\textsuperscript{\rm 3}\footnotemark[2]
}
\affiliations{
\textsuperscript{\rm 1}State Key Laboratory of Robotics and Intelligent Systems, Shenyang Institute of Automation, Chinese Academy of Sciences.\\

\textsuperscript{\rm 2}University of Chinese Academy of Sciences.

\textsuperscript{\rm 3}School of Computing and Data Science, The University of Hong Kong.\\

{\{xiexing, liujiawei, fanhuijie, hanzhi, ytang\}@sia.cn, ziyue\underline{ }lin@connect.hku.hk, liangqqu@hku.hk}}

\usepackage{bibentry}

\begin{document}

\maketitle


\begin{abstract}
We present Autoregressive Representation Alignment (ARRA), a new training framework that unlocks global-coherent text-to-image generation in autoregressive LLMs without architectural modifications. Different from prior works that require complex architectural redesigns, ARRA aligns LLM's hidden states with visual representations from external visual foundational models via a global visual alignment loss and a hybrid token, \texttt{<HYBNEXT>}. This token enforces dual constraints: local next-token prediction and global semantic distillation, enabling LLMs to implicitly learn spatial and contextual coherence while retaining their original autoregressive paradigm. Extensive experiments validate ARRA's plug-and-play versatility. When training T2I LLMs from scratch, ARRA reduces FID by 16.6\% (ImageNet), 12.0\% (LAION-COCO) for autoregressive LLMs like LlamaGen, without modifying original architecture and inference mechanism. For training from text-generation-only LLMs, ARRA reduces FID by 25.5\% (MIMIC-CXR), 8.8\% (DeepEyeNet) for advanced LLMs like Chameleon. For domain adaptation, ARRA aligns general-purpose LLMs with specialized models (e.g., BioMedCLIP), achieving an 18.6\% FID reduction over direct fine-tuning on medical imaging (MIMIC-CXR). These results demonstrate that training objective redesign, rather than architectural modifications, can resolve cross-modal global coherence challenges. ARRA offers a complementary paradigm for advancing autoregressive models. The code is available at \textit{https://github.com/HKU-HealthAI/ARRA}.
\end{abstract}


\section{Introduction}
\label{sec:intro}

\begin{figure*}[t]
    \centering
    \includegraphics[width=0.93\textwidth]{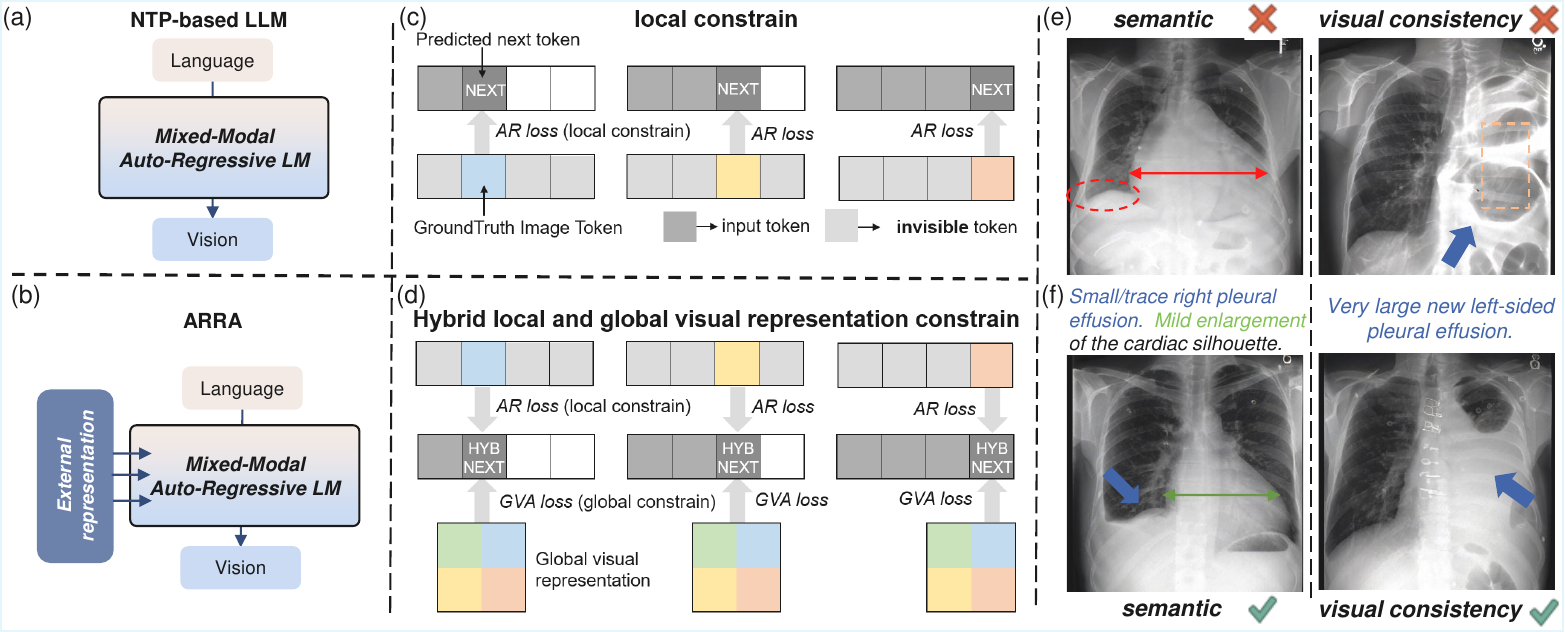}
    \vspace{-2mm}
    \captionof{figure}{ARRA enables high-quality text-to-image generation through a redefined training objective that promotes global coherence. (a)(c) Traditional next-token prediction (NTP)-based LLMs rely solely on the autoregressive loss (AR loss) of the next token $\texttt{<NEXT>}$ for local constraints. (b)(d) ARRA constructs a hybrid token $\texttt{<HYBNEXT>}$, which is aligned by introducing external global visual representation, ensuring that $\texttt{<HYBNEXT>}$ is constrained both locally by the AR loss and globally by the global visual alignment loss (GVA loss). (e)(f) ARRA demonstrates advantages in semantic consistency and visual continuity.}
    \label{fig:first figure}
    \vspace{-3mm}
\end{figure*}

Large language models (LLMs) \cite{achiam2023gpt,yang2025qwen3,guo2025deepseek} and subsequent multimodal LLMs (MLLMs) \cite{team2023gemini,wang2024emu3,liu2024improved,zhu2025internvl3} have revolutionized the field of generative AI. These models, built on the autoregressive (AR) paradigm, show remarkable scalability and generalization capabilities in complex
understanding tasks through a simple yet powerful next-token prediction framework.

Inspired by the success of LLMs, some researchers seek to replicate autoregressive ``next-token prediction" paradigm for text-to-image generation. This paradigm is directly adopted in DALL·E \cite{ramesh2021zero}, Parti \cite{yu2022scalingautoregressivemodelscontentrich}, and LlamaGen \cite{sun2024autoregressivemodelbeatsdiffusion} for image generation, where images are treated as sequences of discrete tokens. However, although ``next-token prediction" paradigm excels in language tasks, where local dependencies naturally align with sequential structure, it struggles to bridge the significant cross-modal gap between language and images. As shown in Fig.~\ref{fig:first figure}(c), optimizing only for local next-token prediction forces the model to focus on isolated token-level features, neglecting the global coherence required for spatially structured visual content.
This sometimes leads to fragmented parts on generated images, such as misaligned ribs in X-rays, where fine-grained details fail to harmonize into a unified whole. It can also cause semantic mismatches, as shown in Fig.~\ref{fig:first figure}(e), where global information is not maintained, leading to inconsistencies in the generated images.

Recognizing this limitation, recent efforts aim to inject global constraints into autoregressive frameworks to fully unlock the potential of LLMs in image generation~\cite{zhou2024transfusion,xie2024show,tian2024visual}. 
Bidirectional attention mechanisms are introduced in Transfusion \cite{zhou2024transfusion} and Show-O \cite{xie2024show} to model global image structure through patch diffusion and mask token modeling, respectively.
These methods achieve promising results in generating high-quality images and demonstrate the potential of LLMs for multimodal generation.
However, they rely on architectural modifications, such as cross-modal attention layers or grafted diffusion modules. While effective, such adaptations often deviate from standard LLM frameworks, limiting their compatibility with pretrained LLMs that excel under pure autoregressive paradigms. For instance, repurposing an off-the-shelf LLM for text-to-image generation would require retraining these modified components, losing benefits of existing scaling laws and generalization capabilities. This practical constraint raises a critical question: \textbf{\textit{Can we unlock the full potential of LLMs for text-to-image generation without altering the original architecture or inference mechanism?}}

\begin{figure*}[t]
    \centering
    \includegraphics[width=0.96\textwidth]{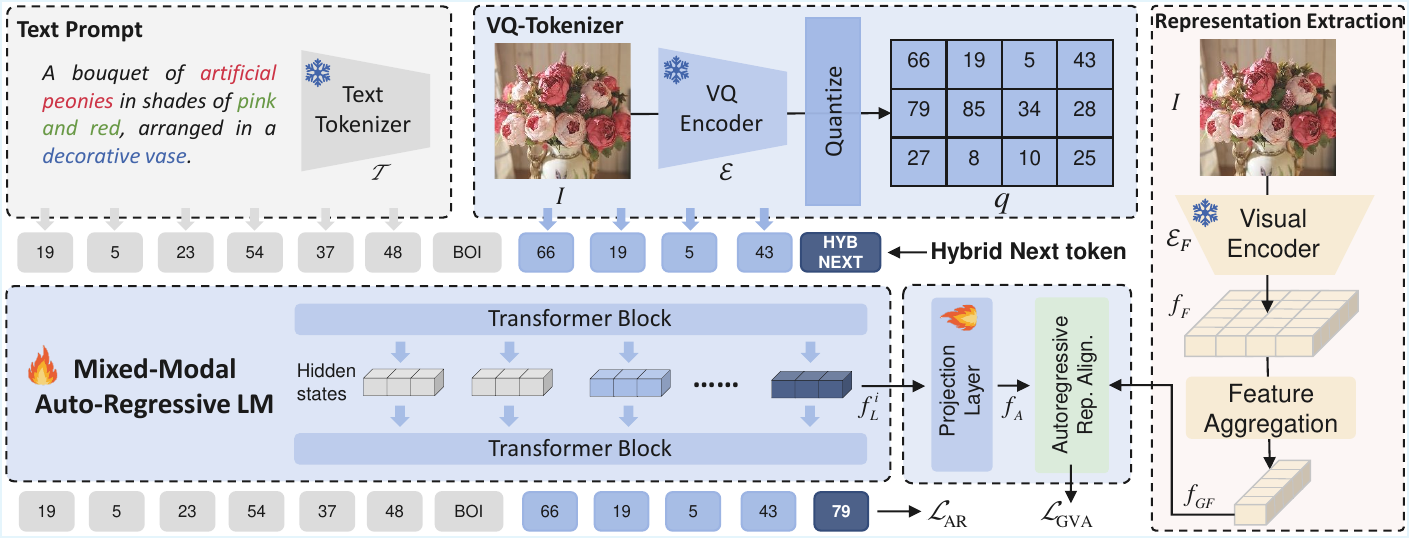}
    \caption{\textbf{Proposed ARRA Framework.} We define the next token predicted in the autoregressive sequence as the ``HYBRID next token'', denoted as $\texttt{<HYBNEXT>}$. During training, $\texttt{<HYBNEXT>}$ is constrained not only \textbf{locally} by the autoregressive loss $\mathcal{L}_{\text{AR}}$ from the next token prediction and LLM codebook matching, but also \textbf{globally} through visual alignment loss $\mathcal{L}_{\text{GVA}}$, which modulates its hidden states using externally well-trained representations. We extract visual representations from a pretrained foundational model and further aggregate these features to obtain semantically enriched representations for alignment.}
    \label{fig:pipline}
\end{figure*}

We address this by proposing Autoregressive Representation Alignment (ARRA), a novel training framework that redefines how LLMs learn text-to-image generation. Different from prior works that modify architectures (e.g., adding attention layers or diffusion modules), ARRA preserves the original LLM framework while injecting global constraints directly into the training objective.
Our key insight is simple: global coherence does not require architectural complexity; it can instead be achieved through a redefined training paradigm.
Specifically, ARRA augments the standard autoregressive loss with a \textbf{global visual alignment loss} that aligns the LLM’s latent representations with semantic guidance from pretrained foundational models (Fig.~\ref{fig:pipline}). To bridge local and global learning, we introduce a hybrid token, \texttt{<HYBNEXT>}, which serves as a bidirectional anchor. Locally, it predicts the next token via standard codebook indices. Globally, its latent embedding aligns with compressed visual features extracted from external models (e.g., BioMedCLIP \cite{zhang2023biomedclip} or MedSAM \cite{MedSAM}) via our novel global visual alignment loss. By distilling rich semantic features (e.g., spatial relationships, object coherence) from external models into the \texttt{<HYBNEXT>} token during training, ARRA enables autoregressive sequences to implicitly learn global structure. More crucially, this alignment occurs only during training, leaving the LLM’s inference process untouched and preserving its inference-time efficiency.

Our experiments demonstrate the versatility of ARRA framework for both natural and medical image generation tasks, achieving improvements without architectural modifications. ARRA supports three key capabilities on AR LLMs:
(1) ARRA enhances training T2I LLMs from scratch. Applied to LlamaGen \cite{sun2024autoregressivemodelbeatsdiffusion} with varying parameter scales, it consistently improves generation performance and exhibits strong scalability.
(2) ARRA effectively transforms pretrained text-generation-only LLMs into T2I generators. When integrated into LLMs without image generation capabilities, such as Chameleon \cite{team2024chameleon}, ARRA yields steady improvements.
(3) ARRA facilitates adaptation of general-purpose generative models to special domains. By integrating domain-specific priors (e.g., BioMedCLIP, MedSAM) into LLMs with image generation capabilities, such as Lumina-mGPT \cite{liu2024lumina}, ARRA substantially outperforms direct fine-tuning.
These capabilities confirm the plug-and-play flexibility of ARRA framework.

The main contributions are summarized below:

(\romannumeral1) 
We propose Autoregressive Representation Alignment, a novel training framework that redefines how LLMs learn text-to-image generation by decoupling global structure learning from model design. By aligning training objective with external representations, ARRA resolves local dependency limitations in LLMs while retaining original architectures and inference efficiency.

(\romannumeral2) 
We introduce the \texttt{<HYBNEXT>} token, a novel mechanism that bridges local next-token prediction with global semantic alignment via distillation from external models (e.g., BioMedCLIP or MedSAM), enabling implicit learning of spatial and contextual relationships.
 
(\romannumeral3) We provide a detailed experimental analysis, offering comprehensive insights into the selection of alignment tokens, aggregation strategy, and external representations. These findings serve as a practical guide for the effective utilization of representation alignment.

(\romannumeral4)
ARRA offers plug-and-play flexibility, facilitating training T2I LLMs from scratch, transforming pretrained text-generation-only LLMs into T2I generators, and adapting general generative models to specific domains, all without architectural modifications. These capabilities are validated on both natural and medical image generation tasks using advanced AR models.


\section{Related Work}
\subsection{Visual Generation Models}

\indent\textbf{Diffusions.} The success of diffusion models has revolutionized the image generation paradigm \cite{qu2015pixel,qu2019wavelet,rombach2022SD,liu2024residual,xie2025dvg}.
 DiTs \cite{peebles2023scalable} show strong scalability by replacing or integrating U-Net with Transformers. This inspires subsequent models such as SD3 \cite{esser2024SD3}, Imagen3 \cite{baldridge2024imagen}, which achieve new state-of-the-art performance in image generation.
Recent work REPA \cite{yu2025repa} explores enhancing diffusion models with external representations.
It utilizes patch-wise representations alignment, matching each patch-level hidden state of the diffusion transformer with corresponding patch tokens from an external encoder.
This alignment improves generation performance without modifying architecture.
However, REPA’s strategy is not directly compatible with AR models. AR models generate image tokens sequentially and do not produce all patch tokens simultaneously during training, making patch-wise alignment infeasible.
Different from this, ARRA introduces a novel hybrid token to bridge local next-token prediction and global alignment, enabling effective integration of the alignment mechanism into AR architectures. For further discussion, please refer to Appendix 2.

\noindent\textbf{Autoregressive models.}  
Early pioneering works, VQ-VAE \cite{van2017neural}, VQ-GAN \cite{esser2021taming} and DALL-E \cite{ramesh2021zero}, demonstrated the potential of AR models for image generation. Subsequent works such as
RQ-Transformer \cite{lee2022autoregressive} also follows a raster-scan manner and enhances image generation performance through extra scales or stacked codes. 
Recent works achieve performance comparable to diffusion models by employing scale modeling \cite{tian2024visual} and eliminating vector quantization \cite{li2024autoregressive,fan2024fluid}.
Additionally, masked prediction autoregressive models \cite{chang2022maskgit,li2024mar} employ BERT-like \cite{devlin2019bert} masked prediction modeling, improving generation efficiency and quality.

\subsection{LLMs for Text-to-Image Generation} 
Currently, researchers pay attention to LLM-based text-to-image generation models, aiming to replicate the success of LLMs in language tasks.
Early works leverage diffusion models as tools to extend LLMs \cite{dong2024dreamllm,ge2024seed,sun2024generative}. These works utilize LLMs as feature extractors to guide diffusion generators for visual generation.
Such models are complex to design and do not fully unleash the generative potential of the LLM for visual generation.
Recent works \cite{sun2024autoregressivemodelbeatsdiffusion,team2024chameleon,lu2024unified,liu2024lumina} attempt to unify text and image modeling within a single LLM, enhancing generation performance through tokenizer optimization \cite{sun2024autoregressivemodelbeatsdiffusion}, early fusion modeling \cite{team2024chameleon}, and flexible resolution modeling \cite{liu2024lumina}. These works discretize both text and images into tokens, which are then fed into the LLM for sequence modeling by next token prediction. However, the local constraints provided by next-token prediction struggle to bridge the cross-modal gap between language and images.

\section{Proposed Method}
We aim to achieve high-quality image generation without altering the next-token prediction paradigm of LLMs. We argue that the inherent cross-domain gaps in LLMs pose significant challenges for image generation tasks when the model lacks the ability to learn global features. To address this, we propose an Autoregressive Representation Alignment (ARRA) framework (see Fig.~\ref{fig:pipline}), which leverages the inherent representation capabilities of well-pretrained foundation models to facilitate the training of complex text-to-image autoregressive generation. Our framework is applied only during training, without affecting inference. 
It can enable the generation of high-quality images with exceptional semantic consistency in a cost-effective manner.

\subsection{Overview}
Our goal is to train an autoregressive model $\mathcal{M}_{\theta}$ leveraging representations derived from an external foundation visual encoder $\mathcal{E}_{F}$. $\mathcal{M}_{\theta}$ takes a text prompt $T$ as input and generates a target image $I$.
During training, $T$ and $I$ are first tokenized into token sequences $s_{T}$ and $s_{I}$, respectively. These token sequences are then used to train the transformer-based autoregressive model $\mathcal{M}_{\theta}$. Meanwhile, foundation visual encoder $\mathcal{E}_{F}$ encodes $I$ into the global visual representation $f_{GF}$, which is used to align with the feature $f_{A}$ extracted from $x_t$ by $\mathcal{M}_{\theta}$. During image generation, the alignment module is removed, and image tokens are generated by $\mathcal{M}_{\theta}$ through next-token prediction. Finally, the output image tokens are decoded into pixel space by an image decoder to produce the target image $I$.
We describe the autoregressive modeling process in Section \ref{Section: Image Generation by Autoregressive Models} and detail our autoregressive representation alignment framework in Section \ref{Section: Visual Representation Alignment}.

\subsection{Modeling via Next-Token Prediction}\label{Section: Image Generation by Autoregressive Models}
The autoregressive architecture comprises two core components: (1) A transformer-based autoregressive model $\mathcal{M_{\theta}}$ for probabilistic modeling of token sequences. (2) A VQ-based model \cite{esser2021taming} with encoder $\mathcal{E}$, quantizer $\mathcal{Q}$ and decoder $\mathcal{D}$ for transformation between image pixels and discrete token sequences.
Formulation involves the following two parts: 

\noindent\textbf{Tokenization.}
In order to apply the next-token prediction modeling in the image domain, it is first required to convert the continuous 2D image pixels into discrete sequences.
This process consists of two steps: (1) 2D image pixels to 2D image tokens, and (2) 2D image tokens to a 1D token sequences.
Specifically, given a image $I\in \mathbb{R}^{H \times W \times 3}$, we first obtain the image feature map $f=\mathcal{E}(I)\in \mathbb{R}^{h\times w \times d}$, where $h=H/c$, $w=W/c$, $d$ is the dimension of the codes, $c$ denotes a compression factor. 
Subsequently, we convert $f$ into discrete tokens by $q = \mathcal{Q}(f) \in Z^{h\times w}$, where the quantizer $\mathcal{Q}(\cdot)$ maps each vector $f^{(i, j)}$ in the image feature map $f$ to the code index $q^{(i, j)}$ of its nearest vector $z^{(i, j)}$ in the codebook $Z$.
The image tokens $q$ are then reshaped into a 1D token sequence  $s_{I} = \{x_1^I, x_2^I, x_3^I, \dots, x_n^I\}$  with a length of $h \cdot w$ , arranged according to the raster scan order. 

For text prompt $T$, we obtain the discrete sequence through $s_{T} = \mathcal{T}(T) = \{x_1^T, x_2^T, x_3^T, \dots, x_n^T\}$, where $\mathcal{T}(\cdot)$ denotes a text tokenizer.

\noindent\textbf{Next token prediction modeling.}
We combine text sequences $s_{T}$ and image sequences $s_{I}$ to obtain discrete tokens  $x = \{x_1, x_2, x_3, \dots, x_n\}$, where $x_n$ is an integer from a tokenizer's vocabulary $V$.
The next-token prediction paradigm posits the probability of current token $x_t$ depends only on its prefix $(x_1, x_2, x_3, \dots, x_{t-1})$. The likelihood of sequence modeling can be expressed as:
\begin{equation}
p ( x ) = \prod _ { t = 1 } ^ { n } p ( x _ { t } | x _ {1}, x _ {2},..., x _ {t-1}).
\end{equation}
The autoregressive model $\mathcal{M_{\theta}}$ formulates the generative task as predicting the distribution of the next token and optimizes the likelihood $p_{\theta}( x )$ through cross-entropy (CE) loss:
\begin{equation}
\mathcal{L}_{\text{AR}}(\theta) = \mathbb{E}_{x_t} [-\log p_{\theta}(x_t | x_{<t})].
\end{equation}

During training, the autoregressive model $\mathcal{M}_{\theta}$ relies on the previous tokens $x_{<t}$ to predict the next token $x_{t}$, where the hidden state of $x_{t}$ in the $i$-th layer of $\mathcal{M}_{\theta}$ is denoted by $f_{L}^i$. In the last layer, the hidden state $f_{L}^{-1}$ is then passed through the LLM head to compute the probability distribution $p_{\theta}$ for $x_{t}$. \textbf{In the original autoregressive model, $p_{\theta}$ is constrained only by the local context of a single token (i.e., $x_t$), lacking the ability to capture global information.} This limitation restricts model's capacity to capture complex cross-modal relationships. To address this problem, we propose Autoregressive Representation Alignment.

\subsection{Autoregressive Representation Alignment}\label{Section: Visual Representation Alignment}
We align the visual representations extracted from the pretrained foundational model with LLM's hidden states and investigate impact of different alignment strategies. The goal of alignment is to enable the hidden state of autoregressive transformer to acquire external global representations, providing meaningful guidance for reconstructing image.

\noindent\textbf{Pre-trained Visual Representation Extraction.} Let $\mathcal{E}_{F}$ be a pretrained foundation model's visual encoder and $I$ be a target image. We encode $I$ as a visual representation by $f_{F} = \mathcal{E}_{F}(I)\in \mathbb{R}^{N\times D}$, where $N,D$ denotes the embedding length and dimension of $f_F$.
$f_{F}$ is aggregated to the global visual representation $f_{GF}$, i.e., $f_{GF} = \text{agg}(f_{F}) \in \mathbb{R}^{1 \times D}$, where $\text{agg}(\cdot)$ denotes a feature aggregation operation. This aggregation operation fully extracts global information in the features and facilitates alignment with the hidden states of autoregressive model. For CLIP series, inspired by \cite{raghu2021vision}, we use the \texttt{<CLS>} token representation from the Transformer-based visual encoder as global visual representation. For SAM series, which lack a \texttt{<CLS>} token, we instead apply average pooling over all patch features for feature aggregation operation. 

\noindent\textbf{Hybrid Next Token.} We define the next token predicted by LLM sequence in our framework as ``\textit{HYBRID next token}'', denoted as $\texttt{<HYBNEXT>}$. Unlike a ``locally constrained token'' in previous autoregressive models that are solely constrained by the LLM codebook, our $\texttt{<HYBNEXT>}$ can fully incorporate external, well-trained global visual representations, making it a ``globally and locally constrained token''.

\noindent\textbf{Global Visual Representation Alignment.}
We obtain the hidden state  $f_{L}^{i}$ of $\texttt{<HYBNEXT>}$ from the autoregressive model $\mathcal{M}_{\theta}$. The hidden state  $f_{L}^{i}$ is converted to $f_{A}\in \mathbb{R}^{1 \times D}$ by a projection layer $\mathcal{A}_{\phi}$ to align with the global visual representation $f_{GF}\in \mathbb{R}^{1 \times D}$, i.e., $f_{A} = \mathcal{A}_{\phi}(f_{L}^{i})$, where $\mathcal{A}_{\phi}$ is a two-layer MLP. 
Representation alignment is achieved through a \textbf{Global Visual Alignment loss} $\mathcal{L}_{\text{GVA}}$, which maximizes the similarity between the projected feature $f_{A}$ and the global visual representation $f_{GF}$:
\begin{equation}
\mathcal{L}_{\text{GVA}}(\theta,\phi) = \text{sim}(f_{A},f_{GF}).
\end{equation}
where \text{sim}($\cdot,\cdot$) denotes cosine similarity loss. This alignment enables the $\texttt{<HYBNEXT>}$ token to learn global visual representation, bridging the cross-modal gap and making the token prediction process more reliable.

Therefore, the autoregressive model can be jointly optimized through the following composite loss function:
\begin{equation}
\mathcal{L}_{\text{ARRA}}(\theta,\phi)= \mathcal{L}_{\text{AR}}(\theta)+\lambda \mathcal{L}_{\text{GVA}}(\theta,\phi).
\end{equation}
$\lambda$ serves as a balancing hyperparameter that controls the relative importance of the alignment objective. Experimentally, we set $\lambda = 1$.





\subsection{Versatile ARRA for Diverse Scenarios} Our ARRA framework is flexible and plug-and-play, supporting different training scenarios and LLM frameworks. Therefore, we provide three representative model variants:

(1) \textbf{ARRA-Base.} It trains an LLM from scratch with random initialization, supporting settings where no pretrained models are available and showcasing ARRA’s ability to learn multimodal alignment from the ground up.

(2) \textbf{ARRA.} It initializes with a pretrained LLM that has strong text generation capabilities, enabling efficient extension to text-to-image tasks with text-generation-only LLMs.

(3) \textbf{ARRA-Adapt.} It builds on pretrained LLMs with both text and image generation capabilities, allowing adaptation to specialized domains such as medical imaging by leveraging special-domain priors.

\section{Experimental Analysis and Results}
We first perform a comprehensive component analysis of the ARRA framework to investigate how different design choices, such as alignment mechanism, feature aggregation strategy, and external encoder selection, impact alignment performance. 
Based on analysis, we finalize the ARRA framework and compare the three model variants, ARRA-Base, ARRA, and ARRA-Adapt, with advanced generative models to evaluate their performance and adaptability.

\subsection{Experimental Setup}
\begin{figure*}[h]
    \centering
        \centering
        \includegraphics[width=0.937\textwidth]{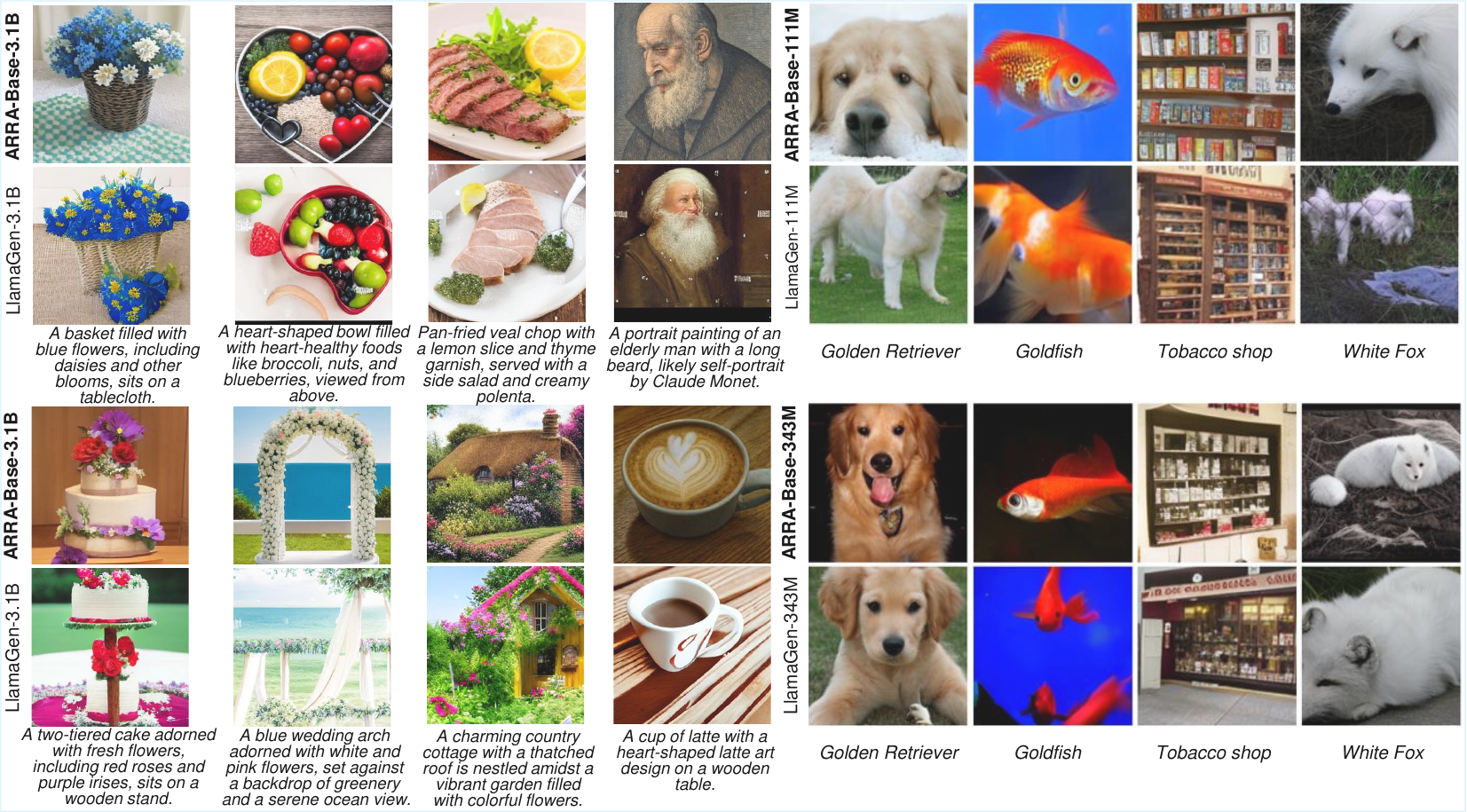}
        \caption{ARRA-Base improves the generation of LlamaGen. Left are text-conditional image generation results on the LAION-COCO dataset, and right are category-conditional image generation results on the Imagenet dataset.}
        \label{fig:laion-coco}
\end{figure*}
\noindent\textbf{Datasets.} 
We evaluate our model on both natural and medical image datasets. For natural images, we conduct evaluations on the ImageNet \cite{imagenet} dataset and a 2.4M high-quality subset of LAION-COCO \cite{schuhmann2022laion}.
For medical imaging, we use the MIMIC-CXR \cite{johnson2019mimic} and DeepEyeNet \cite{Huang2021DeepEyesNet} datasets.
Detailed preprocessing is provided in Appendix 3. 

\noindent\textbf{Implementation Details.} We adopt LlamaGen \cite{sun2024autoregressivemodelbeatsdiffusion}, Chameleon 7B \cite{team2024chameleon} and Lumina-mGPT 7B \cite{liu2024lumina} as respective baseline models for the ARRA-Base, ARRA, and ARRA-Adapt model variants, respectively, to verify the effectiveness of our framework.

\subsection{Component Analysis of Alignment}
\label{Component Analysis of Autoregressive Alignment}

To better understand the core design principles of ARRA and build an effective model, we analyze how different alignment component designs affect the effectiveness of representation alignment in autoregressive generation. 
We focus on the following key questions:
\begin{itemize}
\item  Alignment mechanism: Does token-level alignment (our proposed \texttt{<HYBNEXT>}) outperform fixed-position alignment (\texttt{<REP>}) in preserving global visual constraints during autoregressive generation? (Table \ref{table:alignment_token}) 
\item Feature aggregation strategy: How do different types of features extracted from the same visual encoder affect the generation performance? (Table \ref{table: vit features})
\item  Visual encoder selection on alignment: How does external encoder selection (general vs. specialized and cross-modal vs. vision-only) impact generation? (Table \ref{table:alignment_encoder})
\end{itemize}


\begin{table}[h]
  \centering
  \begin{tabular}{cccc}
    \hline
    Target token& FID $\downarrow$ & MS-SSIM $\uparrow$ & CLIP-Score $\uparrow$ \\\hline
     \rowcolor{gray!20}$\texttt{<REP>}$& 4.85& 0.410& 0.4527\\  
     \rowcolor{gray!20}$\texttt{<HYBNEXT>}$& \textbf{4.15}& \textbf{0.422}& \textbf{0.4576}\\  
    \rowcolor{gray!20} w/o align. & 5.10 & 0.401 & 0.4518 \\  
    \hline
    $\texttt{<REP>}$
& 6.26& 0.401& 0.4499\\  
    $\texttt{<HYBNEXT>}$& \textbf{5.30}& \textbf{0.405}& \textbf{0.4532}\\  
    w/o align. & 7.11 & 0.383 & 0.4460 \\  
    \hline
  \end{tabular}
    \caption{Impact of different alignment mechanism selection on MIMIC-CXR. Gray represents ARRA-adapt model, and white represents ARRA model.} 
    \label{table:alignment_token}
\end{table}



\subsubsection{Alignment mechanism.}
\label{subsec:alignment target token}
We compare two strategies for integrating visual representations: (1) aligning features to a fixed $\texttt{<REP>}$ token at the start of the generated sequence, and (2) aligning to the hidden state of $\texttt{<HYBNEXT>}$, a hybrid token interleaved at every generation step. As shown in Table \ref{table:alignment_token}, $\texttt{<HYBNEXT>}$ yields superior performance. We argue that $\texttt{<HYBNEXT>}$ allows for comprehensive traversal of every token during training sampling, ensuring that each token is effectively constrained by external global representations.
In contrast, $\texttt{<REP>}$ suffers from the ``attention sink'' \cite{liu2024lumina,xiao2023efficient} effect, where attention to the fixed token decays over long sequences, leading to degraded outputs. 
This leads to a key insight: 
\textbf{Takeaway 1.} \textit{Aligning visual representations to a hybrid token interleaved at each generation step $\texttt{<HYBNEXT>}$ is more effective than using a fixed token $\texttt{<REP>}$, as it prevents attention decay and ensures consistent constraint by external representations.}


\begin{table}[h]
  \centering
\begin{tabular}{ccccc}
\hline
   CLS&Avgpool &  FID $\downarrow$ & MS-SSIM $\uparrow$ & CLIP-Score $\uparrow$   
\\ \hline
  \checkmark&& \textbf{5.30}& \textbf{0.405}& \textbf{0.4532}\\ 
  &\checkmark& 6.56& 0.387& 0.4434\\
 \checkmark& \checkmark& 5.93& 0.385&0.4465\\ \hline
\end{tabular}
  \caption{Impact of using different feature aggregation strategies on MIMIC-CXR with ARRA. } 
    \label{table: vit features}
\end{table}

\subsubsection{Feature aggregation strategy.}
\label{subsec:Feature aggregation}
To investigate how different types of features extracted from the visual encoder affect the generation performance, 
we understand three strategies for aggregating features from a single visual encoder: 
(1) the \texttt{[CLS]} token representation, (2) average pooling of all image patch representations, and (3) a concatenation of both representations. 
As shown in Table \ref{table: vit features}, the \texttt{[CLS]} token representation yields optimal performance. We attribute this superiority to the \texttt{[CLS]} token's ability to aggregate global visual information through self-attention mechanisms. This ability provides a compact yet comprehensive representation for cross-modal alignment. This finding consistent with \cite{raghu2021vision}, which establishes that \texttt{[CLS]} tokens capture global representations while patch tokens focus on local features. This global-local distinction suggests that autoregressive image generation models particularly benefit from external global semantic representations as guidance.
These results lead to a key insight: 
\textbf{Takeaway 2.} \textit{The \texttt{[CLS]} token representation in foundation models effectively aggregates global visual information, providing comprehensive guidance for cross-modal alignment.}
\begin{table}[h]
  \centering
  \resizebox{\linewidth}{15mm}{
  \begin{tabular}{cccc}
    \hline
    Target Rep.  & FID $\downarrow$ & MS-SSIM $\uparrow$ & CLIP-Score $\uparrow$ \\\hline
    \rowcolor{gray!20} BioMedCLIP \cite{zhang2023biomedclip} & 4.15 & \textbf{0.422} & \textbf{0.4576} \\  
    \rowcolor{gray!20} Med-SAM \cite{MedSAM}   & \textbf{4.08} & 0.398 & 0.4542 \\  
    \rowcolor{gray!20} CLIP-L \cite{radford2021CLIP}      & 4.63 & 0.407 & 0.4519 \\  
    \rowcolor{gray!20} w/o align. & 5.10 & 0.401 & 0.4518 \\  
    \hline
    BioMedCLIP \cite{zhang2023biomedclip} & 5.30 & \textbf{0.405} & \textbf{0.4532} \\  
    Med-SAM \cite{MedSAM}   & 6.54 & 0.384 & 0.4465 \\  
    CLIP-L \cite{radford2021CLIP}    & \textbf{5.16} & 0.394 & 0.4450 \\  
    w/o align. & 7.11 & 0.383 & 0.4460 \\  
    \hline
  \end{tabular}
  }
   \caption{Impact of alignment with representation extracted from different encoders on MIMIC-CXR.}
    \label{table:alignment_encoder}
\end{table}

\subsubsection{Visual encoder selection on alignment.}
\label{subsec:alignment target representation}
To evaluate how visual encoder selection impacts cross-modal alignment and generation quality, we conduct experiments with three encoders: BioMedCLIP \cite{zhang2023biomedclip} (domain-specific cross-modal encoders), CLIP \cite{radford2021learning} (general-purpose cross-modal encoders), and Med-SAM \cite{MedSAM} (domain-specific pure visual encoders).
As shown in Table~\ref{table:alignment_encoder}, all pretrained encoders improve generation performance compared to without alignment. This improvement arises from global constraints imposed by external representations, which regularize the autoregressive generation process. 
Notably, for training from a text-generation-only LLM (ARRA), BioMedCLIP and CLIP demonstrate superior performance. Their cross-modal training helps bridge the gap between text and image modalities, enabling LLMs to learn what to generate (semantics) before how to generate (pixel details). Conversely, when fine-tuning pretrained LLMs with image-generation capabilities (ARRA-Adapt), domain-specific encoders like BioMedCLIP and MedSAM dominate. BioMedCLIP injects medical-specific semantics, while MedSAM provides structural priors (e.g., organ shapes) through its segmentation-focused features. These results lead to a key insight: 
\textbf{Takeaway 3.} \textit{When the LLM lacks image generation capabilities, cross-modal encoders are crucial for semantic grounding. However, for LLMs with image generation capabilities, domain-specific encoders are more effective, as they provide fine-grained features needed for domain-specific adaptation.}

\subsection{Main Comparison Results} 
\label{subsec:comparison_results}
\begin{figure}[h]
    \centering
    \includegraphics[width=\linewidth]{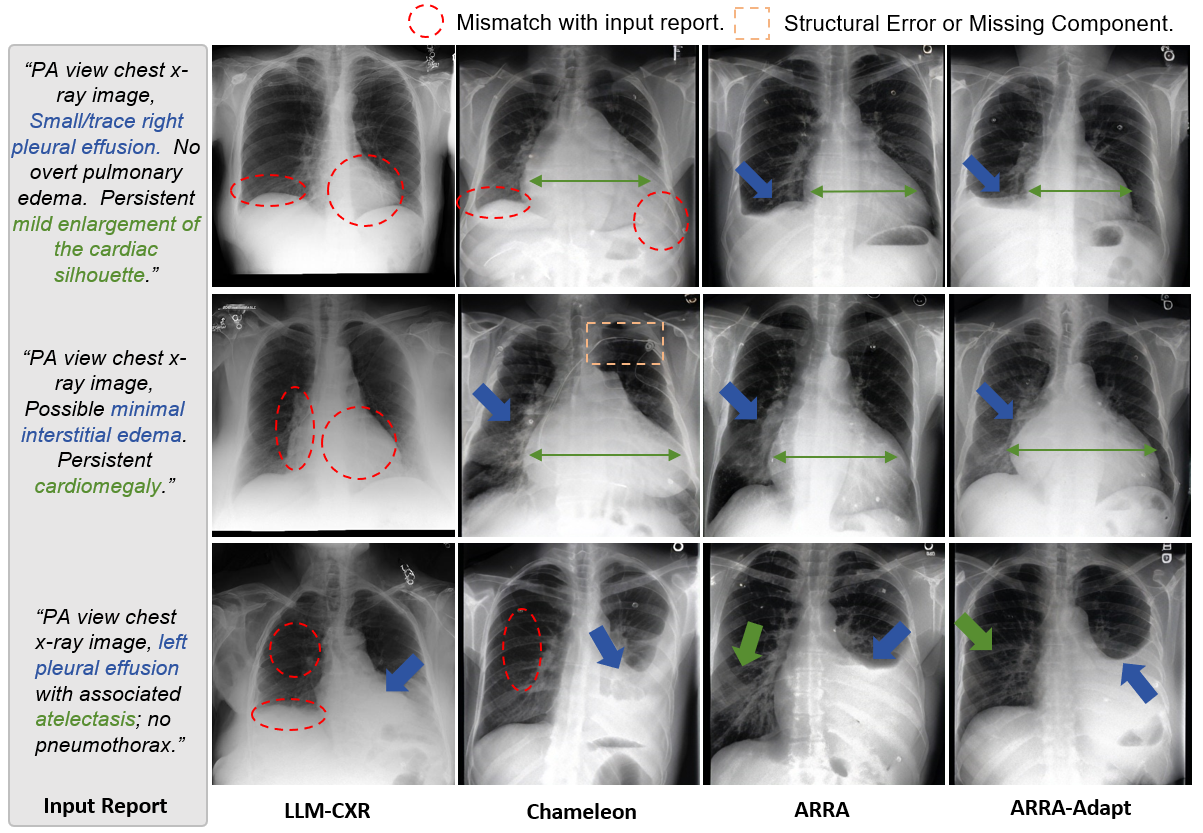}
    \caption{Visual comparison of different methods with ARRA and ARRA-Adapt on the MIMIC-CXR dataset.}
    \label{fig:results_xray}
\end{figure}
Based on analysis, we finalize ARRA framework with CLIP and BioMedCLIP for alignment on natural and medical datasets, respectively. We use \texttt{[HYBNEXT]} and \texttt{[CLS]} as our alignment strategy. We compare ARRA framework under the three proposed model variants (ARRA-Base, ARRA, ARRA-Adapt) to demonstrate its superiority and versatility.

\begin{table*}[h]
  \centering
  \resizebox{\linewidth}{!}{
  \begin{tabular}{lcc|cc||lcccc}
    \toprule
    \multicolumn{5}{c||}{\textbf{ImageNet}} & \multicolumn{5}{c}{\textbf{LAION-COCO}} \\
    \cmidrule(r){1-5} \cmidrule(l){6-10}
    \multirow{2}{*}{Model} & \multicolumn{2}{c|}{111M} & \multicolumn{2}{c||}{343M} & \multirow{2}{*}{Model} & 343M & 775M & 1.4B & 3.1B \\
    & FID ↓ & IS ↑ & FID ↓ & IS ↑ & & (FID, CLIP-Score) & (FID, CLIP-Score) & (FID, CLIP-Score) & (FID, CLIP-Score) \\
    \midrule
    LlamaGen     & 5.47          & 193.6        & 4.33& \textbf{286.6}& LlamaGen     & (12.78, 0.2470)& (11.79, 0.2512)& (11.57, 0.2510)& (11.88, 0.2605)\\
    ARRA-Base    & \textbf{5.08} & \textbf{201.2} & \textbf{3.61}& 262.6& ARRA-Base    & (\textbf{11.67}, \textbf{0.2500})& (\textbf{10.81}, \textbf{0.2571})& (\textbf{10.50}, \textbf{0.2589})& (\textbf{10.45}, \textbf{0.2615})\\
    \bottomrule
  \end{tabular}
  }
    \caption{Quantitative performance comparison of ARRA-Base and LlamaGen models with different scales on ImageNet datasets and 2.4M subset of LAION-COCO datasets, respectively.}
    \label{table:imagenet and laion-coco}
\end{table*}

\begin{table*}[h]
  \centering
  \small  
  \setlength{\tabcolsep}{3pt}  
\begin{tabular}{ccccc|ccc}  
\hline
 & \multicolumn{4}{c}{Chest-Xray: MIMIC-CXR} & \multicolumn{3}{c}{Fundus: DeepEyeNet} \\  
\hline
 Method &  FID $\downarrow$ & IS $\uparrow$ & MS-SSIM $\uparrow$ & CLIP-Score $\uparrow$   
        & FID $\downarrow$ &   MS-SSIM $\uparrow$ & CLIP-Score $\uparrow$   
        \\  
\hline
 Original SD V2-1\cite{rombach2022high} & 71.68 & 2.365 & 0.128 & 0.2333 
        & 166.45  & 0.141 & 0.2164 
       \\  
 DreamBooth SD \cite{ruiz2023dreambooth} & 60.40 & 2.269 & 0.270 & 0.3693 
        & 80.70  & 0.352 & 0.3061 
   \\ 
  MINIM \cite{wang2025self} & 15.62 & 2.468 & 0.317 & 0.4423 
        & 59.71  & 0.333 & 0.2862 
        \\  
 UniXGen \cite{lee2024vision} & 30.75 & 2.437 & 0.361 & 0.4128 
        & -  & - & - 
        \\  

 LLM-CXR \cite{lee2024llm} & 5.88 & 2.134 & 0.395 & 0.4374 
        & -  & - & - 
   \\ 
 Chameleon \cite{team2024chameleon} & 7.11 & 2.498 & 0.383 & 0.4460 
        & 38.37  & 0.341 & 0.3234 
   \\ 
\cline{1-8}
 \textbf{ARRA} & 5.30 & 2.587 & 0.405 & 0.4532 
        & 35.01  & 0.376 & 0.3354 
        \\  
 \textbf{ARRA-Adapt} & \textbf{4.15} & \textbf{2.746} & \textbf{0.422} & \textbf{0.4576} 
        & \textbf{34.70}  & \textbf{0.392} & \textbf{0.3389} 
        \\  
\hline
\end{tabular}
  \caption{Compare different methods with ARRA and ARRA-Adapt on MIMIC-CXR and DeepEyeNet datasets, respectively.}
    \label{table: comparison sota}
\end{table*}

\noindent\textbf{Facilitating training T2I LLMs from scratch: performance comparison of ARRA-Base.} 
We evaluate the generation performance and model scalability of ARRA-Base variant, which is trained from scratch, on large-scale datasets. We first conduct experiments on ImageNet, using LlamaGen \cite{sun2024autoregressivemodelbeatsdiffusion} with 111M and 343M parameters as baselines. We then train models on a 2.4M high-quality subset of LAION-COCO and employ LlamaGen variants of increasing sizes (343M to 3.1B parameters) as baselines. As shown in Table \ref{table:imagenet and laion-coco} and Fig. \ref{fig:laion-coco}, ARRA-Base outperforms the baseline model across all scales. The FID decreased by 16.6\% and 12.0\% on the ImageNet and LAION-COCO datasets, respectively.
Meanwhile, as the model size increased, the performance remained steadily improved (FID decreased from 11.67 to 10.45),  demonstrating ARRA maintains strong scalability of autoregressive model. 
These results lead to a key insight:
\textbf{Takeaway 4.} \textit{\textbf{ARRA-Base} enables efficient training of T2I LLMs from scratch, while preserving strong model scalability.}

\noindent\textbf{Boosting pretrained LLMs for T2I generation and domain adaptation: performance of ARRA and ARRA-Adapt.} 
We evaluate ARRA and ARRA-Adapt, initialized with text-generation-only and general image generation pretrained LLMs, respectively, on the MIMIC-CXR and DeepEyeNet datasets.
Both variants are benchmarked against state-of-the-art diffusion models (Stable Diffusion \cite{rombach2022high}, DreamBooth \cite{ruiz2023dreambooth}, MINIM \cite{wang2025self}) and autoregressive models (LLM-CXR \cite{lee2024llm}, UniXGen \cite{lee2024vision}, Chameleon \cite{team2024chameleon}). 
As shown in Table \ref{table: comparison sota}, our model achieves superior performance in both visual quality and semantic alignment.
Compared to the chameleon, ARRA achieves a 25.5\% and 8.8\% reduction in FID on the MIMIC-CXR and DeepEyeNet datasets, respectively.
On the MIMIC-CXR dataset, ARRA surpasses the best-performing baseline, LLM-CXR, reducing the FID from 5.88 to 5.30 and increasing the CLIP-Score from 0.4374 to 0.4532. The ARRA-Adapt variant delivers even stronger results, achieving an FID of 4.15 and a CLIP-Score of 0.4576.
In addition, as shown in Fig. \ref{fig:results_xray}, our models demonstrate superior alignment with fine-grained clinical details, such as lesion location and severity.
These results lead to a key insight: 
\textbf{Takeaway 5.} \textit{\textbf{ARRA}  enables a more effective transformation of text-generated-only LLMs into T2I generators, while \textbf{ARRA-Adapt} substantially improves domain adaptation and aligns general image-generation LLMs with specialized fields more effectively, both outperforming baseline approaches.}

\vspace{2.5mm}
\section{Conclusion}
We propose Autoregressive Representation Alignment (ARRA), a framework that enhances autoregressive image generation by injecting external visual representations during training. This approach enriches global semantic understanding while maintaining the model’s original autoregressive paradigm during generation. Experiments on natural and medical image generation tasks demonstrate ARRA’s versatility, offering a cost-effective framework for training autoregressive text-to-image generation models. Our work bridges the gap between multimodal domains and provides novel insights into modeling unified multimodal generation.
\clearpage

\section{Acknowledgments}
This work was supported by the National Natural Science Foundation of China (62306253, 61873259), the Early Career Fund (27207025), the National Key Research and Development Program of China under Grant 2024YFB4707700, the National Natural Science Foundation of China under Grant U23A20343 and the Guangdong Natural Science Fund-General Program (2024A1515010233).

\bibliography{aaai2026}
\clearpage

\makeatletter
\@ifundefined{isChecklistMainFile}{
  \newif\ifreproStandalone
  \reproStandalonetrue
}{
  \newif\ifreproStandalone
  \reproStandalonefalse
}
\makeatother

\ifreproStandalone
\documentclass[letterpaper]{article} 
\usepackage[submission]{aaai2026}  
\usepackage{times}  
\usepackage{helvet}  
\usepackage{courier}  
\usepackage[hyphens]{url}  
\usepackage{graphicx} 
\urlstyle{rm} 
\def\UrlFont{\rm}  
\usepackage{natbib}  
\usepackage{caption} 
\frenchspacing  
\setlength{\pdfpagewidth}{8.5in} 
\setlength{\pdfpageheight}{11in} 
%
\usepackage{algorithm}
\usepackage{algorithmic}
\usepackage{tcolorbox}
%
\usepackage{newfloat}
\usepackage{listings}

\usepackage{epsfig}
\usepackage{amsmath}
\usepackage{amssymb}
\usepackage{pifont}
\usepackage{makecell}
\usepackage{booktabs}
\usepackage{multirow}

\usepackage{silence}

\makeatletter

\usepackage{colortbl}  
\usepackage{xcolor} 

\DeclareCaptionStyle{ruled}{labelfont=normalfont,labelsep=colon,strut=off} 

\lstset{%
	basicstyle={\footnotesize\ttfamily},
	numbers=left,numberstyle=\footnotesize,xleftmargin=2em,
	aboveskip=0pt,belowskip=0pt,%
	showstringspaces=false,tabsize=2,breaklines=true}
\floatstyle{ruled}
\newfloat{listing}{tb}{lst}{}
\floatname{listing}{Listing}
%
\pdfinfo{
/TemplateVersion (2026.1)
}

\setcounter{secnumdepth}{2} 

%


\title{Unleashing the Potential of Large Language Models for Text-to-Image Generation through Autoregressive Representation Alignment

Supplementary Material}

\usepackage{bibentry}

\begin{document}
\maketitle
\fi

\setcounter{page}{1}
\setcounter{table}{0}
\setcounter{section}{0}
\setcounter{figure}{0}

\label{sec: More visualizations on MIMIC-CXR}
\begin{figure*}[h]
    \centering
    \includegraphics[width=\textwidth]{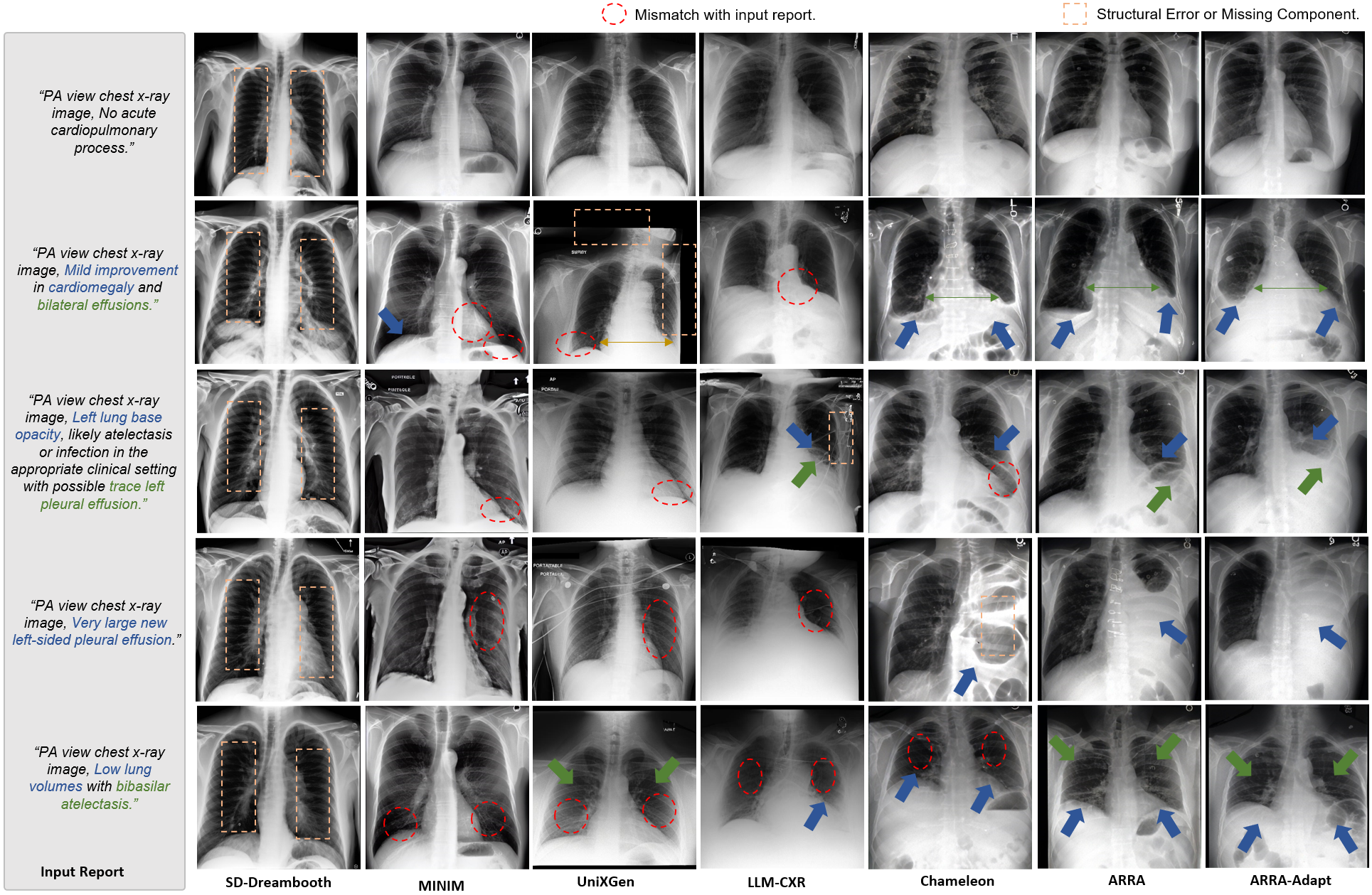}
    \caption{More visualizations on the chest X-ray generation task across the MIMIC-CXR datasets.}
    \label{fig:addition_results_xray}
    \vspace{-2mm}
\end{figure*}
\section{Overview}
Our supplementary materials include the following sections:
\begin{itemize}
\item Discussion with related work (Section~\ref{sec: Discussion with related work}).
    \item Additional implementation details about ARRA and other comparative methods in MIMIC-CXR and DeepEyeNet (Section~\ref{sec: More Implementation Details});
    \item Implementation details of component analysis of alignment. (Section~\ref{sec:Implementation details of component analysis.});
    \item Ablation study, including generalizability at different resolutions, choices of projection layers, regularization coefficients $\lambda$, training objectives and alignment depth (Section~\ref{sec: Ablation study});
    \item More visual results on the MIMIC-CXR dataset (Section~\ref{sec:More visualizations on MIMIC-CXR});
    \item Implementation details and more results on the LAION-COCO and ImageNet datasets(Section~\ref{sec:Implementation details and visualization of ImageNet datasets});
\end{itemize}

\section{Discussion with related work}
\label{sec: Discussion with related work}
\noindent\textbf{Discussion with REPA \cite{yu2025repa}}: While both works leverage feature alignment concepts, ARRA presents fundamentally distinct contributions from REPA:

\noindent\textbf{Different motivation:} REPA resolves challenges in learning high-quality internal representations for diffusion models by injecting representations from external clean images. In sharp contrast, ARRA addresses a fundamental limitation of autoregressive LLMs in text-to-image generation: the inherent inability to capture global coherence.  Unlike prior works requiring complex architectural redesigns to impose global constraints, ARRA demonstrates that such global coherence naturally emerges through our alignment-driven training paradigm, achieved by aligning LLM hidden states with visual representations from an external visual encoder. 

\noindent\textbf{\bf REPA not compatible with autoregressive models (AR): } REPA utilizes patch-wise alignment, matching each patch-level hidden states of the diffusion transformer with the patch token from an external encoder, i.e., {\bf all patches of DiTs $\Leftrightarrow$ all patches of external encoder}. But AR does not output all tokens of image patches during training, i.e., {\bf wo/ all patches of AR $\nLeftrightarrow$ all patches of external encoder}. Thus, REPA is incompatible with AR.
To make the feature alignment idea compatible with the design of AR, we thus design a novel hybrid token \texttt{<HYBNEXT>}, which serves as a bidirectional anchor: local next token prediction via standard codebook and global information injection via a feature alignment loss. Results in Sec. 4.2.1 validate the superiority of our \texttt{<HYBNEXT>} design.

\section{Additional implementation details in MIMIC-CXR and DeepEyeNet}
\label{sec: More Implementation Details}

\noindent\textbf{Datasets.} For the \textbf{MIMIC-CXR}, we employ three distinct radiographic views: posteroanterior (PA), anteroposterior (AP), and lateral (LATENT). Input prompts are generated from the ``impression'' section of diagnostic reports, formatted as: ``\{view\} view chest X-ray image, \{impression\}''. We select 221,238 pairs for training and 1,000 pairs for testing. \textbf{DeepEyeNet} \cite{Huang2021DeepEyesNet} is a comprehensive fundus dataset, containing more than 15k high-quality “text-image” pairs. In our experiments, we derive prompts from the ``clinical-description'' section and selectively choose 7,190 training pairs and 1,089 testing pairs. All images are preprocessed through center cropping to 512×512 pixels.

\noindent\textbf{Evaluation Metrics.}
 We employ three quantitative metrics to evaluate the proposed method: Fréchet Inception Distance (FID), Multi-Scale Structural Similarity (MS-SSIM), and CLIP-Score. For X-ray image generation, FID is computed using features extracted from a CXR-pre-trained DenseNet-121 (XRV) \cite{cohen2022torchxrayvision}, while for fundus images, we utilize features from an ImageNet-pre-trained Inception V3 \cite{heusel2017gans}. The CLIP-Score metric is derived from the cosine similarity between image-text feature pairs encoded through BioMedCLIP \cite{zhang2023biomedclip}, a large-scale medical vision-language model.

\noindent\textbf{ARRA and ARRA-Adapt.} All experiments on medical images are implemented using PyTorch on 4 NVIDIA A6000 Ada GPUs. We employ a batch size of 6 with a learning rate of 2e-5, optimized by AdamW (weight decay=0.05, $\beta_{1}$=0.9, $\beta_{2}$=0.95). To stabilize training, we apply z-loss regularization \cite{team2024chameleon} with a weight of 1e-5. All data are pre-tokenized before training to increase throughput. The VQ tokenizer operates with a downsampling rate of 16, resulting in a 1024-token representation for each image. 

\noindent\textbf{Original SD V2-1. \cite{rombach2022high}} Stable-diffusion-2-1 is a diffusion model trained from natural images, and we obtain its official pre-training weights, which are used directly to generate images without training, to demonstrate its zero-shot generation capability.

\noindent\textbf{DreamBooth SD. \cite{ruiz2023dreambooth}} We adapt Dreambooth to fine-tune the stable-diffusion-xl-base model, choosing “X-rays” and “fundus” as class identifiers [class noun] on the MIMIC-CXR and DeepEyesNet datasets, respectively, and supervised the fine-tuning using the full data.

\noindent\textbf{MINIM. \cite{wang2025self}} MINIM is a medical image generation model based on stable-diffusion-v1-4 for full fine-tuning. We use its official implementation for training and generation.

\noindent\textbf{UniXGen. \cite{lee2024vision}} The UniXGen model is a unified model for X-ray image understanding and generation, fully trained on the MIMIC-CXR dataset. We use its official implementation and obtain its pre-training weights for generation.

\noindent\textbf{LLM-CXR. \cite{lee2024llm}} The LLM-CXR model is a unified model for X-ray image understanding and generation using dolly-v2-3b \cite{DatabricksBlog2023DollyV2} as a baseline for full parameter fine-tuning training. We implement training and inference on the MIMIC-CXR dataset using its official implementation.

\noindent\textbf{Chameleon. \cite{team2024chameleon}} Chameleon is a multimodal large language model trained on the natural image datasets. Its open-source weights are only for visual language understanding. We follow its official implementation and use 7B pre-training weights to fine-tune full parameters for training on the x-ray and fundus image dataset.

\section{Implementation of component analysis}
\label{sec:Implementation details of component analysis.}

\noindent For the \textbf{Alignment mechanism}, we align the representations extracted from BioMedCLIP.

\noindent For the \textbf{Feature aggregation strategy}, we align the representations extracted from the BioMedCLIP model to the hidden state of $\texttt{<HYBNEXT>}$.

\noindent For the \textbf{Visual encoder selection on alignment}, we align the representations to the hidden state of $\texttt{<HYBNEXT>}$.

\section{More Ablation study}
\label{sec: Ablation study}


\subsubsection{Generalizability at different resolutions.}
We validate the generalization capability of the ARRA framework at resolutions of 256×256 and 512×512, as shown in Table \ref{table: Generalizability at different resolutions}. Experimental results demonstrate that ARRA can stably improve the generation performance of baseline models at different resolutions, demonstrating its resolution generalization ability.
\begin{table}[h]
  \centering
  \caption{Generalizability at different resolutions.}
  \label{table: Generalizability at different resolutions}
\resizebox{\linewidth}{10mm}{
\begin{tabular}{ccccc}
\hline
  Resolution&Model&  FID $\downarrow$ & MS-SSIM $\uparrow$ & CLIP-Score $\uparrow$   
\\ \hline
 256×256&Chameleon& 9.42& 0.369& 0.4405\\
 256×256&ARRA& \textbf{6.11}& \textbf{0.388}&\textbf{0.4436}\\\hline
 512×512& Chameleon
& 7.11& 0.383&0.4460\\
 512×512& ARRA& \textbf{5.30}& \textbf{0.405}&\textbf{0.4532}\\\hline
\end{tabular}
  }
\end{table}

\begin{table}[h]
  \centering
  \caption{Compare performance of different alignment heads.}
  \label{table: alignment head}
\begin{tabular}{ccccc}
\hline
   MLP&Maxpool&  FID $\downarrow$ & MS-SSIM $\uparrow$ & CLIP-Score $\uparrow$   
\\ \hline
  \checkmark&& \textbf{5.30}& \textbf{0.405}& \textbf{0.4532}\\ 
  &\checkmark& 5.37& 0.386& 0.4375\\\hline
\end{tabular}
\vspace{-2mm}
\end{table}

\subsubsection{Projection layer.}
We evaluate the performance of different projection layers, including a two-layer MLP and a direct pooling layer. As shown in Table \ref{table: alignment head}, our experiments reveal that using a two-layer MLP outperforms the pooling layer, even when the projection layer is discarded during testing.
We hypothesize that, compared to the non-trainable pooling layer, employing a trainable MLP as a "soft link" during training allows the model to better learn information-rich and precise representations while adaptively filtering out irrelevant information.

\subsubsection{Effect of $\lambda$.}
We also examine the effect of the regularization coefficient $\lambda$ by training models with different coefficients from 0.5 to 2 and comparing the performance. As shown in Table \ref{table: loss coefficient}, the optimal value is reached when the regularization coefficient $\lambda$ = 1.
\begin{table}[h]
  \centering
  \caption{Ablation study for alignment loss coefficient $\lambda$.}
  \label{table: loss coefficient}
\begin{tabular}{cccc}
\hline
  $\lambda$&  FID $\downarrow$ & MS-SSIM $\uparrow$ & CLIP-Score $\uparrow$ \\ \hline
 0.5& 6.22& 0.389& 0.4468\\ 
 0.8& 5.38& 0.399& 0.4460\\
 1& \textbf{5.30}& \textbf{0.405}&\textbf{0.4532}\\
 1.5& 5.87& 0.394&0.4477\\
 2& 6.12& 0.390&0.4479\\\hline
\end{tabular}
  \vspace{-2mm}
\end{table}

\subsubsection{Training objective.}
We compare two alignment objectives: MSE loss and cosine similarity loss. As shown in Table \ref{table: loss objective}, both achieve similar FID and MS-SSIM scores, but cosine similarity yields better semantic accuracy (higher CLIP-Score). This advantage stems from its focus on directional similarity rather than absolute distances, making it more robust for cross-modal alignment tasks.

\begin{table}[h]
  \centering
  \caption{Impact of different alignment objectives.}
  \label{table: loss objective}
\begin{tabular}{cccc}
\hline
  Objective&  FID $\downarrow$ & MS-SSIM $\uparrow$ & CLIP-Score $\uparrow$ \\ \hline
 Cos. sim.& \textbf{5.30}& \textbf{0.405}& \textbf{0.4532}\\
 MSE& 5.32& 0.405&0.4454\\\hline
\end{tabular}
\end{table}




\subsubsection{Alignment depth.}
\label{subsec:alignment depth}
As shown in Table \ref{table:alignment_layer}, we investigate the effects of representation alignment at different layers of the Transformer architecture. Our results demonstrate that alignment at various layers consistently enhances model performance, with early-layer alignment (e.g., layer 1) yielding optimal results. We argue that early-layer alignment provides sufficient global semantic guidance while allowing subsequent layers to focus on capturing high-frequency details, thereby establishing robust representations for image generation. 

These results lead to a key insight:
\textbf{Takeaway 4.} \textit{Performing representation alignment at earlier LLM layers provides sufficient global semantic guidance while allowing deeper layers to focus on capturing high-frequency details.}
\begin{table}[h]
  \centering

  \caption{Impact of aligning representation to different layers on MIMIC-CXR. Gray part represents the ARRA-adapt, and the white part represents the ARRA.} 
  \label{table:alignment_layer}
  \begin{tabular}{lccc}
    \hline
    Layer Index & FID $\downarrow$ & MS-SSIM $\uparrow$ & CLIP-Score $\uparrow$ \\ \hline
    \rowcolor{gray!20} layer1  & \textbf{4.15} & \textbf{0.422} & 0.4576 \\  
    \rowcolor{gray!20} layer5  & 4.58 & 0.421 & 0.4587 \\  
    \rowcolor{gray!20} layer20 & 4.46 & 0.413 & 0.4563 \\  
    \rowcolor{gray!20} layer30 & 4.51 & 0.412 & 0.4537 \\  
    \rowcolor{gray!20} layer31 & 4.53 & 0.418 & \textbf{0.4606} \\  
    \rowcolor{gray!20} w/o align. & 5.10 & 0.401 & 0.4518 \\  
    \hline
    layer1  & 5.30  & \textbf{0.405} & \textbf{0.4532} \\  
    layer5  & 6.21  & 0.385 & 0.4467 \\  
    layer20 & 5.70  & 0.404 & 0.4500 \\  
    layer30 & \textbf{5.04}  & 0.398 & 0.4478 \\  
    layer31 & 5.71  & 0.373 & 0.4500 \\  
    w/o align. & 7.11  & 0.383 & 0.4460 \\  
    \hline
  \end{tabular}
\end{table}



\begin{figure}[h]
    \centering
    \includegraphics[width=\linewidth]{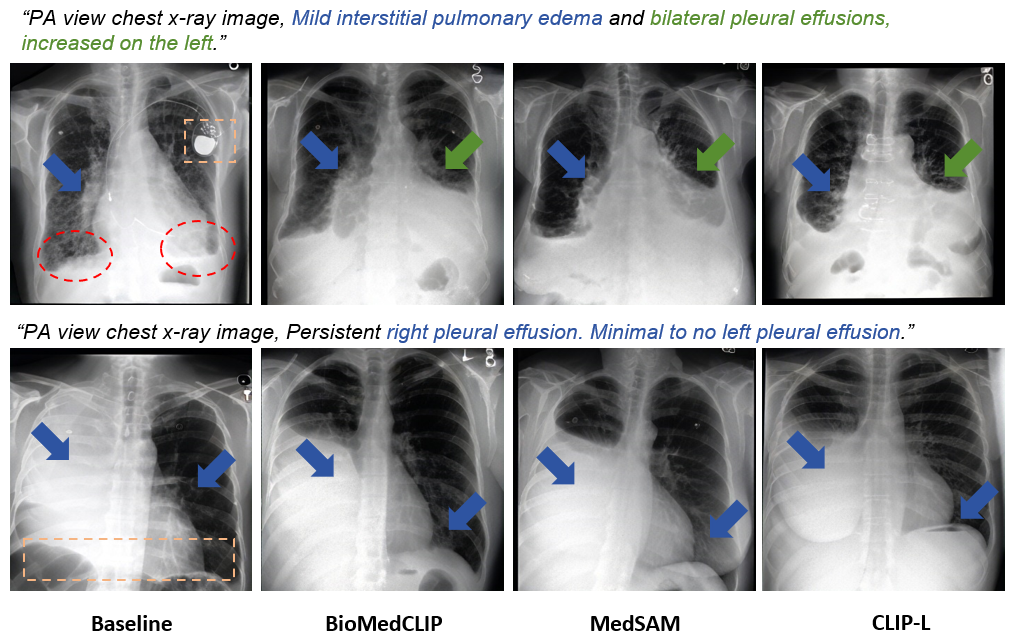}
    \caption{Visualization of alignment with features extracted from different encoders.}
    \label{fig:encoder ablation}
    \vspace{-2mm}
\end{figure}


\begin{figure}[h]
    \centering
    \includegraphics[width=\linewidth]{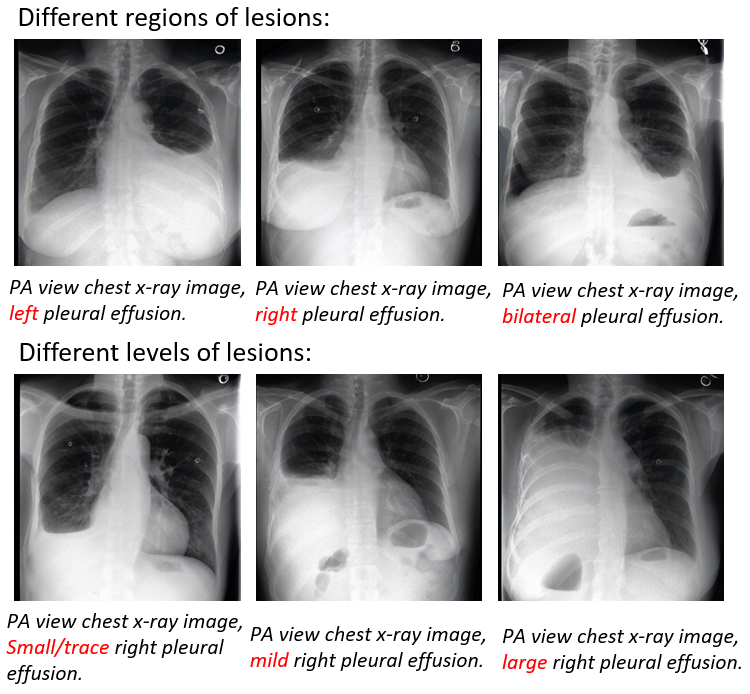}
    \caption{Visualization of different lesion localizations and lesion levels.}
    \label{fig: different lesions}
\end{figure}

\begin{figure}[h]
    \centering
    \includegraphics[width=\linewidth]{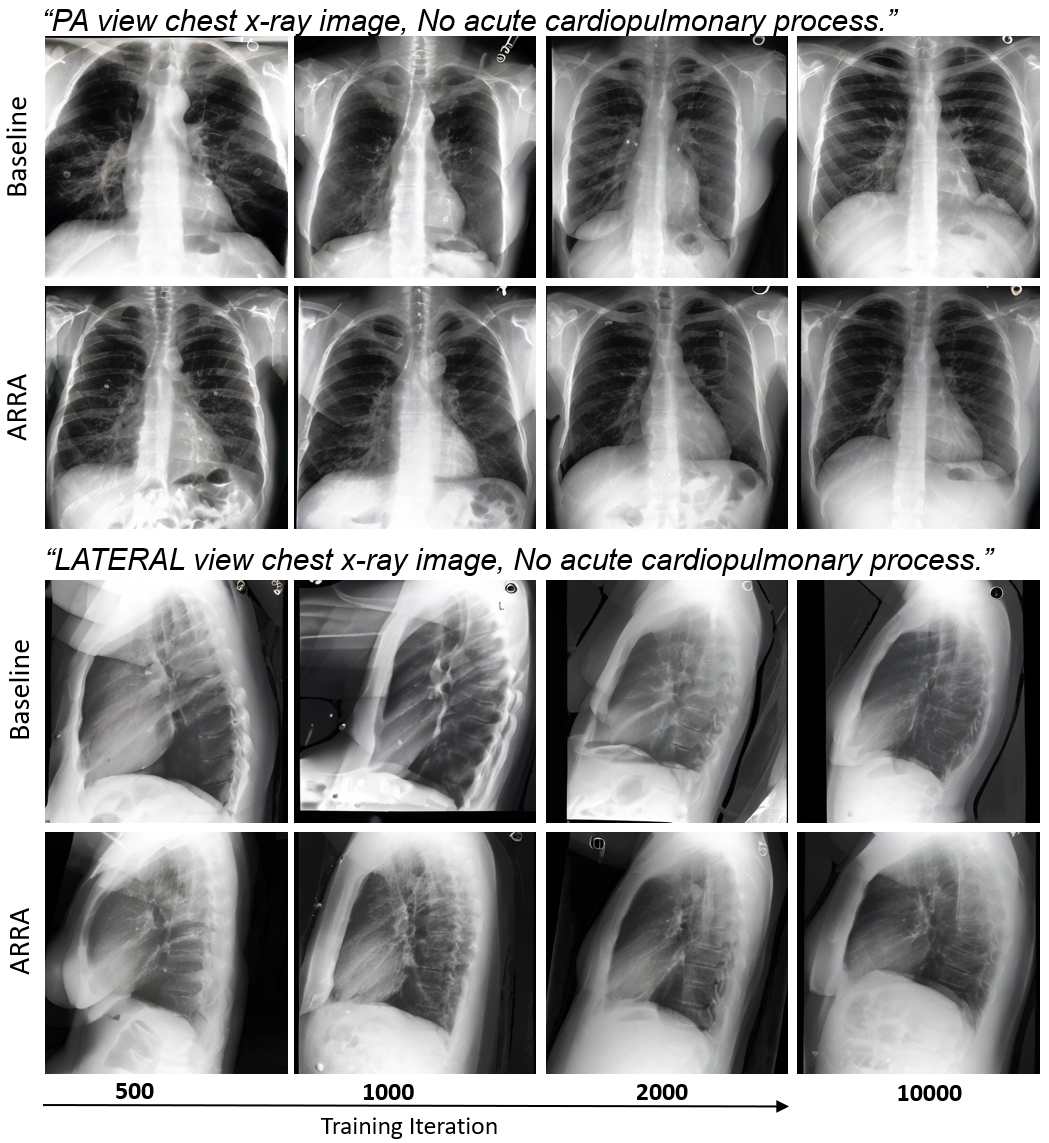}
    \caption{More visual comparison of images generated by the baseline and the ARRA model (with alignment) during the iteration process.}
    \label{fig:iteration_small_sup}
    \vspace{-2mm}
\end{figure}

\section{More visualizations on MIMIC-CXR}
\label{sec:More visualizations on MIMIC-CXR}
We present additional visual results from experiments on MIMIC-CXR, including:
\begin{itemize}
    \item Visual comparisons of ARRA with other methods on X-ray image generation tasks (Fig.~\ref{fig:addition_results_xray});
    \item Visualizations of baseline models using different pretrained encoders for alignment (Fig.~\ref{fig:encoder ablation});
    \item Visualization of different lesion localizations and lesion levels. (Fig.~\ref{fig: different lesions});
    \item Convergence performance visualizations of baseline models and those using the ARRA framework (Fig.~\ref{fig:iteration_small_sup}).
\end{itemize}

\section{More implementation and results in LAION-COCO and ImageNet}
\label{sec:Implementation details and visualization of ImageNet datasets}
\subsection{Datasets}
We collect 2.4M high-quality image-text pairs from the \textbf{LAION-COCO} \cite{schuhmann2022laion} dataset as our training data. We additionally collect 10K samples as a test set. All images are center-cropped to 256×256. We use subset provied by GoT \cite{fang2025got}, filtering with similarity $>$ 0.3.
For \textbf{ImageNet} \cite{imagenet}, we conduct experiments on the ImageNet 256×256 conditional generation benchmark.

\subsection{Implementation details}
\noindent\textbf{ARRA-Base.} We apply the ARRA framework on LlamaGen \cite{sun2024autoregressivemodelbeatsdiffusion} to test the generalization ability of our model. In our experiments, we use a batch size of 32 with a learning rate of 1e-4 and a vocabulary size of 16,384. For the visual foundation model, we use CLIP-L \cite{radford2021CLIP}, where the extracted feature, our target representation, has the shape (batch\_size, 512). In the 111M model, the dimension of the hidden states is 768, while in the 343M model, the dimension of the hidden states is 1024. We use an MLP layer as the feature summarization tool to reduce the dimension of hidden states, making it possible to be aligned with our target. We set $\lambda = 1$ and use cosine similarity to calculate the loss between the target representation and the hidden states.

\begin{figure*}[h]
    \centering
    \includegraphics[width=\textwidth]{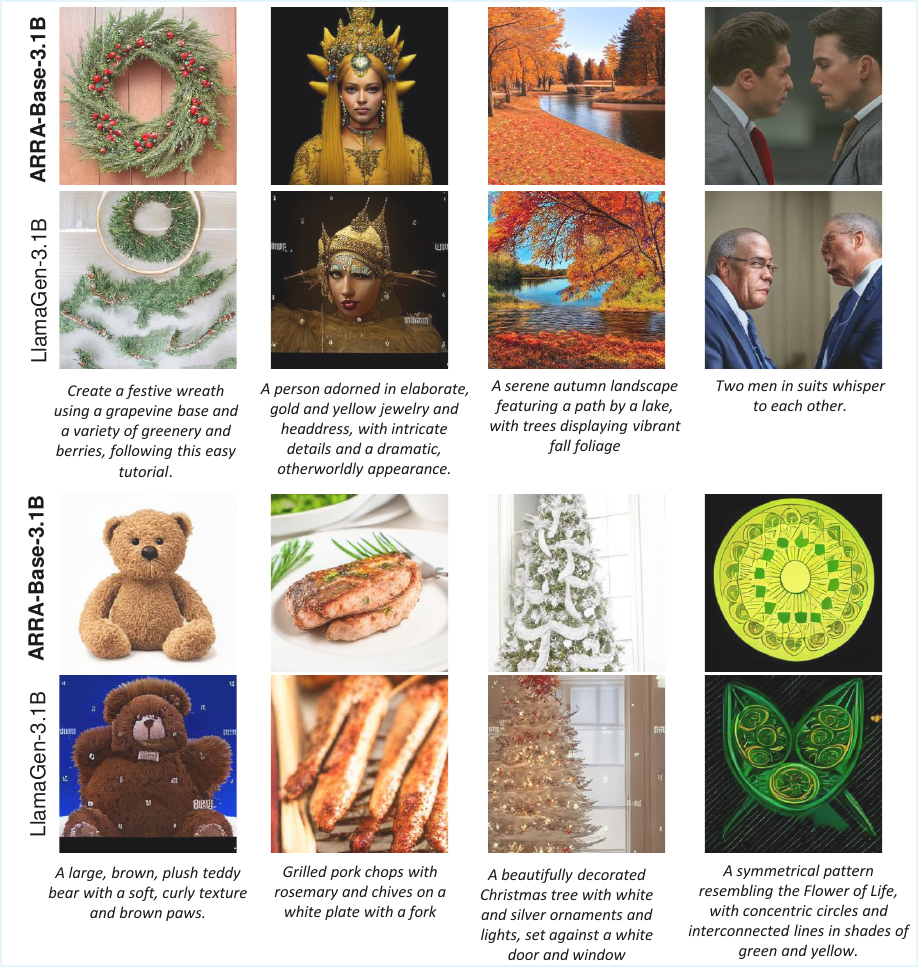}
    \caption{More visualizations of ARRA-Base-3.1B on the LAION-COCO dataset.}
    \label{fig:addition_coco_results}
\end{figure*}

\begin{table}[h]
  \centering
  \caption{ARRA-Base in image generation task across the ImageNet Dataset (classifier-free guidance=2.00)}
  \label{table: ARRA full}
  \begin{tabular}{c c c c | c c}  %
  \toprule
 Method & Para & epochs & data & FID $\downarrow$ & IS $\uparrow$ \\ 
 \midrule
  \rowcolor{gray!20} LlamaGen & 111M & 100  & 0.26M  & 9.328  & 126.4   \\
 \rowcolor{gray!20} LlamaGen & 111M & 150  & 0.26M  & 8.421  & 142.7   \\
 \rowcolor{gray!20}  LlamaGen & 111M & 200  & 0.26M  & 8.010  & 145.4   \\
  \rowcolor{gray!20} LlamaGen & 111M & 250  & 0.26M  & 7.580  & 161.8\\ 
  \rowcolor{gray!20} LlamaGen & 111M & 300  & 0.26M  & 7.283  & 156.7   \\ 
  \rowcolor{gray!20}LlamaGen & 111M & 300  & 1.28M  & 5.464  & 193.6   \\
  
    LlamaGen & 343M & 100  & 0.26M  & 5.192  & 187.8   \\
    
 LlamaGen & 343M & 150  & 0.26M  & 4.666  & 206.7   \\
 LlamaGen & 343M & 200  & 0.26M  & 4.419  & 202.0    \\
  LlamaGen & 343M & 300  & 0.26M  & 4.294  & 218.6\\ 
LlamaGen & 343M & 300  & 1.28M   & 4.327  & 286.6   \\

 \midrule
  \rowcolor{gray!20}ARRA-Base & 111M & 100  & 0.26M  & 8.710  & 133.0   \\
\rowcolor{gray!20} ARRA-Base & 111M & 150  & 0.26M  & 7.873  & 142.2   \\ 
\rowcolor{gray!20}ARRA-Base & 111M & 200  & 0.26M  & 7.483  & 145.8   \\
\rowcolor{gray!20}ARRA-Base & 111M & 250  & 0.26M  & 6.737  & 162.7   \\
\rowcolor{gray!20}ARRA-Base & 111M & 300  & 0.26M  & 6.653  & 159.0\\ 

ARRA-Base & 343M & 100  & 0.26M  & 4.982  & 192.1   \\
ARRA-Base & 343M & 150  & 0.26M  & 4.544  & 200.8   \\ 
ARRA-Base & 343M & 200  & 0.26M  & 4.182  & 206.8   \\
ARRA-Base & 343M & 300  & 0.26M  & 3.971  & 219.1   \\ 
 
 \bottomrule
\end{tabular}
\end{table}

\subsection{More experiments results}
This is a more comprehensive experiment results of ARRA-Base on the Imagenet datasets (Table \ref{table: ARRA full}), where we prove that ARRA enhances the generation ability of LlamaGen. ARRA works for both 111M and 343M models, indicating that this method is scalable.

\subsection{Visualizations}
\begin{figure}[h]
    \centering
    \includegraphics[width=\linewidth]{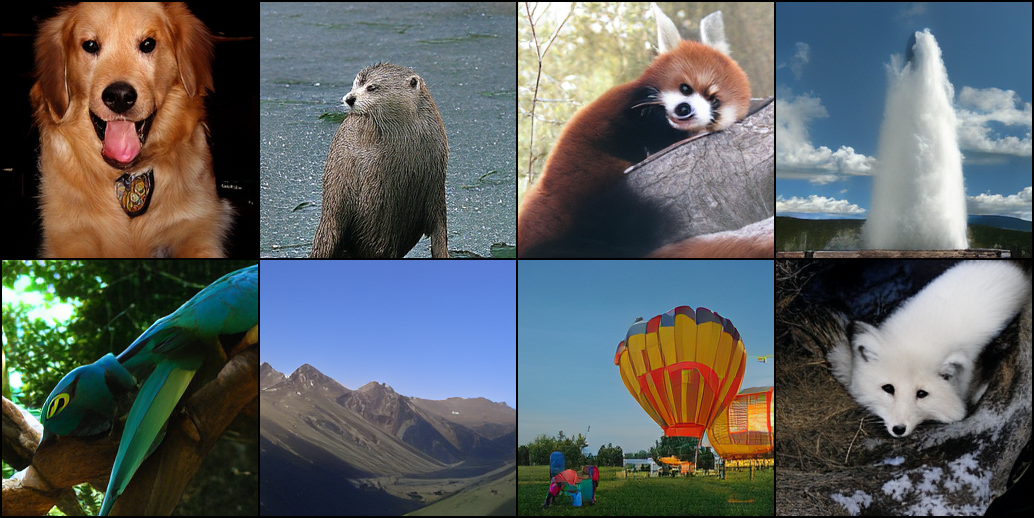}
    \caption{More visualizations of ARRA-Base-343M on ImageNet dataset.}
    \label{fig:addition_results_natural2}
\end{figure}

\begin{figure}[h]
    \centering
    \includegraphics[width=\linewidth]{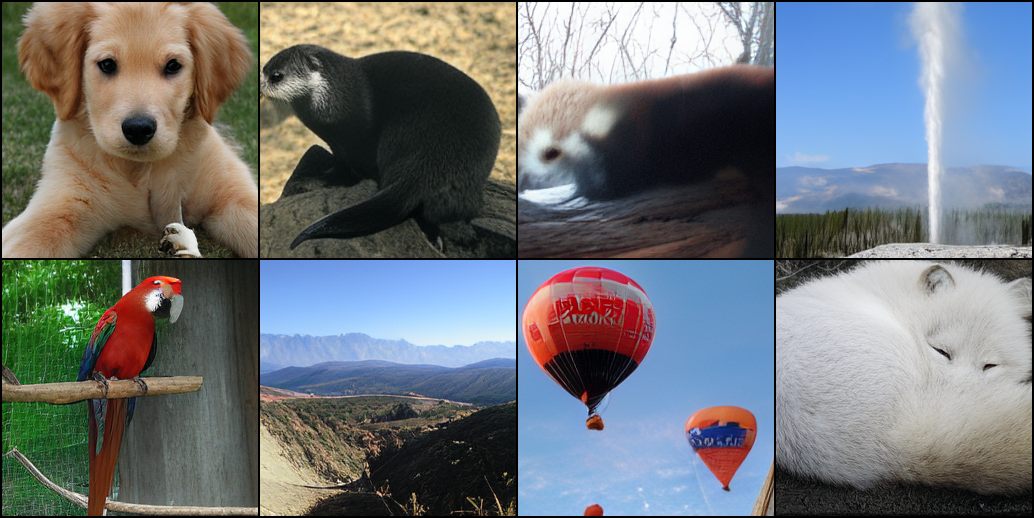}
    \caption{More visualizations of LlamaGen-343M on ImageNet dataset.}
    \label{fig:addition_results_natural1}
\end{figure}

\begin{figure}[h]
    \centering
    \includegraphics[width=\linewidth]{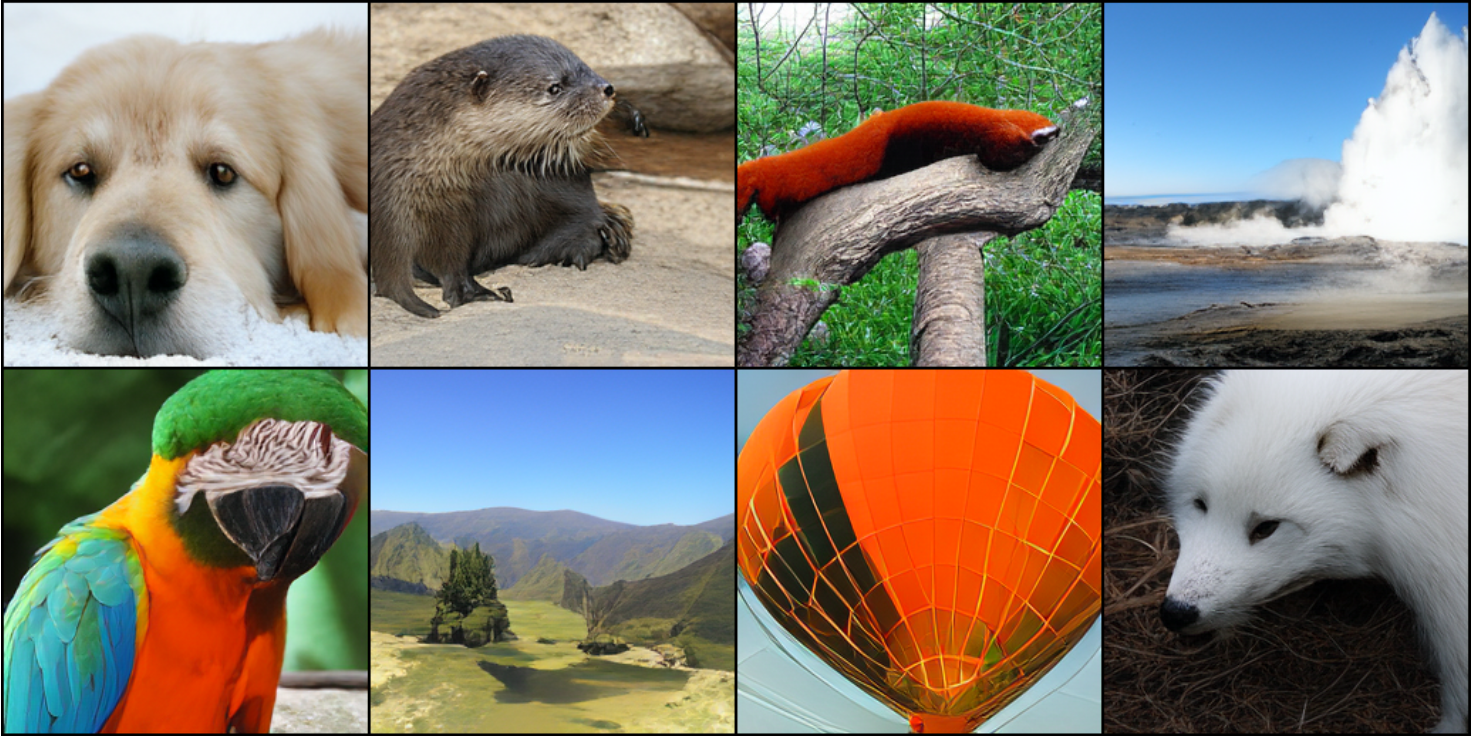}
    \caption{More visualizations of ARRA-Base-111M on ImageNet dataset.}
    \label{fig:addition_results_natural4}
\end{figure}

\begin{figure}[h]
    \centering
    \includegraphics[width=\linewidth]{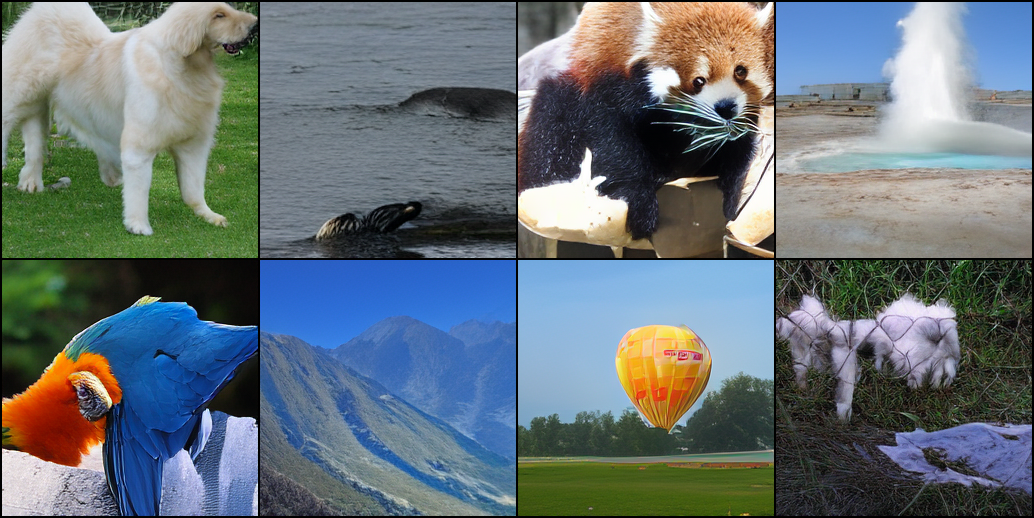}
    \caption{More visualizations of LlamaGen-111M on ImageNet dataset.}
    \label{fig:addition_results_natural3}
\end{figure}

In this section, we present additional visualizations (Fig. 
\ref{fig:addition_coco_results},
\ref{fig:addition_results_natural2},
\ref{fig:addition_results_natural1},
\ref{fig:addition_results_natural4},
\ref{fig:addition_results_natural3}) on the LAION-COCO and ImageNet datasets, comparing ARRA-Base and LlamaGen.
\label{sec: More visualizations on Natural Images}

\clearpage

\end{document}
\end{document}